\PassOptionsToPackage{hyperfootnotes=false}{hyperref}
\PassOptionsToPackage{table}{xcolor}
\documentclass[11pt]{article}

\usepackage[final]{acl}

\usepackage{times}
\usepackage{latexsym}
\usepackage{microtype}
\usepackage{graphicx}
\usepackage{subcaption}
\usepackage{booktabs} %
\usepackage{multirow}
\usepackage{cuted}
\usepackage[table]{xcolor}
\usepackage{amsmath}
\usepackage{amssymb}
\usepackage{mathtools}
\usepackage{amsthm}
\usepackage{enumitem}
\usepackage{marvosym}
\usepackage{bbm}
\usepackage{listings}
\usepackage{booktabs}
\usepackage[normalem]{ulem}

\usepackage[most]{tcolorbox}
\usepackage{adjustbox}
\usepackage{booktabs}
\usepackage{xspace}
\usepackage[most]{tcolorbox}
\usepackage{enumitem} %
\usepackage[capitalize,noabbrev]{cleveref}
\crefname{section}{Sec.}{Secs.}
\Crefname{section}{Appendix}{Appendices}
\crefname{table}{Table}{Tables}
\crefname{figure}{Fig.}{Figs.}
\crefname{algocf}{alg.}{algs.}
\Crefname{algocf}{Algorithm}{Algorithms}

\theoremstyle{plain}

\theoremstyle{definition}

\theoremstyle{remark}

\usepackage{tabularx}
\newcommand{\nbf}[1]{{\noindent \textbf{#1}}}

\usepackage[T1]{fontenc}

\usepackage[utf8]{inputenc}

\usepackage{microtype}

\usepackage{inconsolata}

\usepackage{graphicx}

\title{The Target Is a Predicate: Predicate-State Grounding for Semantic-Level Video Reasoning Segmentation}

\newcommand{\bench}{\texttt{PS-Bench}\xspace}
\newcommand{\method}{PSG\xspace}

\author{
\textbf{Yuhuan Hong}\textsuperscript{1}\thanks{Equal contribution.} \quad
\textbf{Yiming Ren}\textsuperscript{1}\footnotemark[1] \quad
\textbf{Zhiqiang Lin}\textsuperscript{1}\footnotemark[1] \quad
\textbf{Feize Wu}\textsuperscript{2}\footnotemark[1] \quad
\textbf{Xingyun Liu}\textsuperscript{3}
\\
\textbf{Yitong Xie}\textsuperscript{3} \quad
\textbf{Lei Zhou}\textsuperscript{4} \quad
\textbf{Lisheng Wu}\textsuperscript{4} \quad
\textbf{Junjie Wang}\textsuperscript{1}\footnotemark[2] \quad
\textbf{Yujiu Yang}\textsuperscript{1}\thanks{Corresponding author.}
\\[2pt]
\textsuperscript{1}Tsinghua University \quad
\textsuperscript{2}Peking University \\
\textsuperscript{3}Sun Yat-sen University \quad
\textsuperscript{4}Yinwang intelligent Technology Co., Ltd
\\[2pt]
\texttt{hongyh25@mails.tsinghua.edu.cn} \quad
\texttt{yang.yujiu@sz.tsinghua.edu.cn}
}

\begin{document}
\maketitle

\begin{abstract}

In this paper, we introduce Predicate-State Video Reasoning Segmentation (PS-VRS), where a query defines a dynamic predicate $P(o,t)$ over objects and time rather than a fixed object identity. Existing methods typically treat the target as permanently active once identified, leading to over‑emission of masks during vacant intervals and failure to handle target switching. To address this, we construct \bench, a large‑scale benchmark with 2,077 videos and 16,250 queries annotated with explicit active/vacant predicate states, temporal intervals, and pixel masks. We further propose Predicate-State Grounding (PSG), a State‑Bind‑Emit pipeline that: (1) predicts \texttt{[ACTIVE]/[VACANT]} state tokens; (2) aligns them to video frames via dense cross‑modal similarity; and (3) emits masks via SAM2 only for active states. Experiments show that PSG outperforms existing baselines by $+9.32$ $\mathcal{J}\&\mathcal{F}$ on \bench and significantly reduces false‑positive mask emissions. Our results demonstrate the effectiveness of explicit predicate-state modeling for video reasoning segmentation with dynamic visual witnesses.
\end{abstract}

\begin{figure*}[t]
\centering
\includegraphics[width=0.98\textwidth]{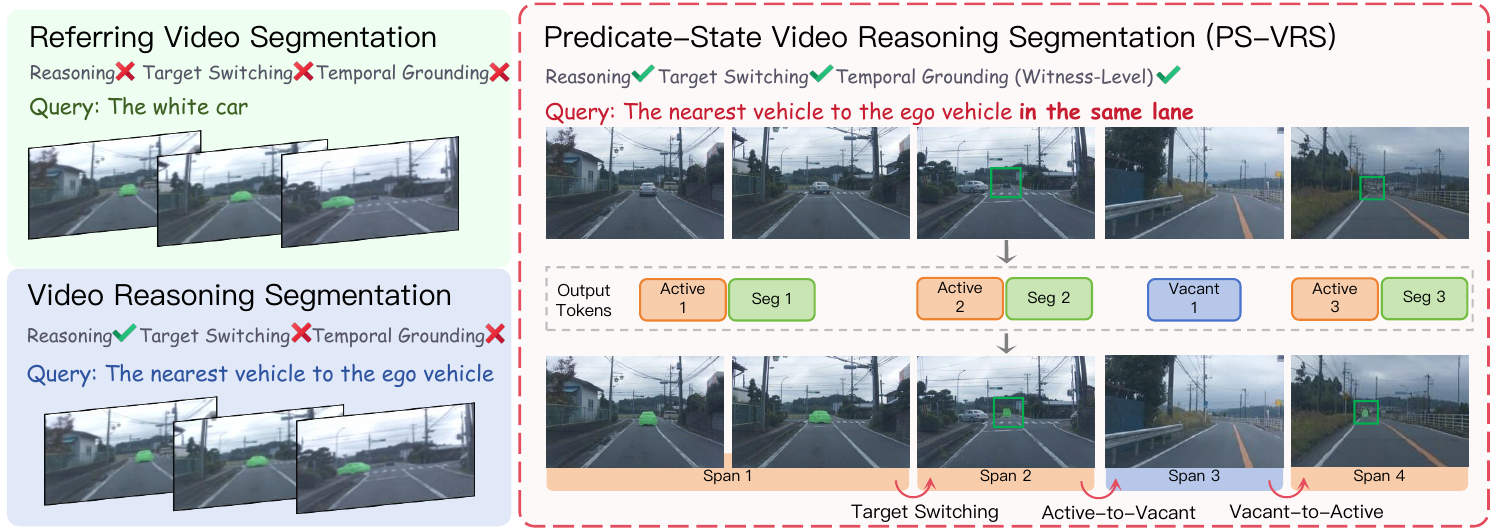}
\caption{
Task formulation comparison.
Referring Video Object Segmentation segments explicitly referred targets, while Video Reasoning Segmentation introduces implicit reasoning queries but typically lacks explicit target switching and temporal predicate-state boundaries.
PS-VRS treats the query as a predicate over objects and time, requiring reasoning, target switching, active/vacant temporal grounding, and pixel-level masks for active segments.
}
\label{fig:placeholder}
\vspace{-14pt}
\end{figure*}

\section{Introduction}

Language‑guided video segmentation has evolved from referring to explicit objects towards reasoning over implicit conditions, relations, and temporal dynamics~\citep{lai2024lisa, VideoLISA,shen2025reasonseg_survey,shen2025survey}. In many real‑world scenarios, a user query does not specify a persistent object identity but instead defines a predicate over objects and time. For example, “If the person releases his right hand, what object will fall?” — the valid visual witness may be an apple at one moment, become absent for several frames, and later become a tissue. The query defines a time‑dependent Boolean predicate $P_Q(o,t)$: at each moment, a model must decide whether any object satisfies the predicate, segment the current visual witness if it exists, and emit no mask otherwise.

As shown in~\cref{fig:placeholder}, existing task formulations cover only part of this setting. 
Referring Video Object Segmentation~\citep{wu2022referformer,mevis} assumes a fixed target object referred to by the query. Video Reasoning Segmentation~\citep{GROUNDMORE,visa} introduces implicit queries that require world knowledge or motion understanding. However, their annotation and evaluation protocols typically treat the target as an active referent once identified, without explicitly representing predicate‑state transitions such as target switching, active‑to‑vacant, or vacant‑to‑active changes. Consequently, they do not directly evaluate whether a visible object remains query‑valid, or whether the predicate has no valid witness at a given moment. In fact, many VRS methods couple semantic validity with mask generation, predicting what to segment without explicitly deciding whether the query predicate is active~\citep{GLUS,VideoLISA}. This results in severe over‑emission of invalid masks, especially during vacant intervals or when the target identity switches.

To address this, we introduce Predicate-State Video Reasoning Segmentation (PS-VRS).
Under PS-VRS, a reasoning query is formalized as $P_Q(o,t)$, and the output must satisfy three requirements:
1) \textbf{Predicate-state reasoning}, determining whether the predicate is active or vacant over time;
2) \textbf{Witness-level temporal grounding}, localizing semantic segments and their transition boundaries;
and 3) \textbf{State-conditioned pixel segmentation}, segmenting the current visual witness only for active states while emitting no mask for vacant states.

Then, we propose PSG (Predicate‑State Grounding), a method that decouples the task via a State‑Bind‑Emit mechanism. PSG first predicts the active/vacant predicate state using dedicated \texttt{[ACTIVE]} and \texttt{[VACANT]} tokens, associates active states with their current visual witnesses via dense cross‑modal alignment, and conditionally emits masks only under active states using \texttt{[SEG]} tokens projected into a SAM2~\citep{ravi2024sam2segmentimages} decoder. This design separates semantic validity from mask generation, supporting different visual witnesses across active segments while reducing the emission of invalid masks in vacant states.

To evaluate more complex language-guided segmentation scenarios, we construct PS‑Bench, a large‑scale benchmark with 2,077 videos, 16,250 queries, and approximately $1.57$M masks. Each query is annotated as a sequence of temporal predicate states with active/vacant labels, target descriptions, reasoning annotations, and pixel‑level masks for active intervals. \bench has an average duration of $45.81$ seconds and $3.82$ semantic segments per sample, including $2.18$ active and $1.64$ vacant segments, making it substantially more challenging than existing datasets.
Compared with existing benchmarks, \bench jointly evaluates implicit reasoning, dynamic visual witnesses, predicate-state temporal grounding, and pixel-level segmentation.

Experiments show that PSG significantly outperforms strong baselines on PS‑Bench, achieving a $+9.32$ $\mathcal{J}\&\mathcal{F}$ improvement and reducing false‑positive mask emissions in vacant states by a large margin. Ablations confirm that explicit state modeling is critical for this improvement.

Our contributions are threefold:
\begin{itemize}[nosep,leftmargin=*]
\item We define Predicate-State Video Reasoning Segmentation (PS-VRS) and introduce \bench, a benchmark for evaluating dynamic visual witnesses, active/vacant predicate states, visible-but-invalid cases, and pixel-level masks in videos.
\item We propose \method, a Predicate-State Grounding method that decomposes segmentation into predicate-state inference, witness binding, and state-conditioned mask emission.
\item We provide a diagnostic evaluation showing that active-frame metrics understate full-video grounding failures, while explicit predicate-state modeling reduces invalid mask emission under query-conditioned transitions.
\end{itemize}

\section{Related Work}

\nbf{From Referring to Predicate-State Reasoning Video Segmentation.}~Task formulations in language-guided video segmentation have evolved from referring video object segmentation (RVOS) to reasoning video segmentation (RVS). This progression aligns with broader efforts to make multimodal outputs more controllable and their supporting evidence more traceable~\citep{ren2025anycap,ren2026sin}. RVOS datasets such as Ref‑YouTube‑VOS ~\citep{ref-youtube-vos}, Ref‑DAVIS~\citep{ref-davis}, and MeViS ~\citep{mevis} assume that the query describes a persistent object that can be tracked across frames. The model’s output is a continuous mask track for that object. MeViS introduced motion expressions (``the person walking slowly”), but still assumes a single, continuously present target.

Reasoning segmentation~\citep{GROUNDMORE, lai2024lisa, li2025revseg, kao2026cotrvs, munasinghe2025videoglamm} extends this to implicit queries that require world knowledge, causal inference, or relational reasoning. In the video domain, recent works such as VISA~\citep{visa}, VideoLISA~\citep{VideoLISA}, VRS‑HQ~\citep{VRS-HQ}, VideoGLaMM~\citep{munasinghe2025videoglamm},  GLUS~\citep{GLUS}, and VideoSeg‑R1~\citep{VideoSeg-R1} combine MLLMs with mask decoders to generate spatiotemporal masks for queries that do not name a specific object~\citep{bagchi2025refereverything, zhang2025video, video-of-thought, zheng2025villa, han2025videoespresso}. However, their annotation and evaluation protocols still treat the target as an active referent once identified; they do not explicitly represent predicate states where the valid witness may disappear (active → vacant) or switch to a different object (target switching). Consequently, these methods tend to over‑emit masks in vacant intervals and fail to adapt to dynamic predicate transitions.

\nbf{State Modeling in Video Segmentation.}%
~Modeling temporal state changes has been explored in action segmentation, event detection, and object state tracking~\citep{bagchi2025refereverything, han2025videoespresso}. However, most work focuses on predefined state vocabularies (e.g., “open/closed”) rather than query‑defined predicates. Closer to our setting, \cite{video-of-thought} and \cite{zheng2025villa} incorporate chain‑of‑thought reasoning into video segmentation, but they still lack explicit vacant state modeling. To the best of our knowledge, no prior benchmark or method formalizes the predicate as a time‑dependent Boolean function over objects, requiring the model to output both active and vacant intervals with conditional mask generation.

\nbf{Benchmark Comparison.}
Table~\ref{tab:stats} summarizes existing language‑guided video segmentation benchmarks. 
While MeViS~\citep{mevis} and GroundMore~\citep{GROUNDMORE} provide some temporal diversity, they do not annotate vacant segments or target switching. \bench introduces three novel features: (i) explicit active/vacant predicate‑state labels; (ii) an average of $3.82$ semantic segments per sample ($2.18$ active, $1.64$ vacant); and (iii) long videos ($45.81$ s average), requiring the model to reason about state changes over extended durations. To the best of our knowledge, PS-Bench is the first benchmark designed specifically for predicate‑state video reasoning segmentation.

\begin{table}[!t]
\centering
\small
\setlength{\tabcolsep}{3.5pt}
\renewcommand{\arraystretch}{1.05}
\resizebox{\linewidth}{!}{
\begin{tabular}{lccccc}
\toprule
Benchmark & Videos & Dur./Vid. & Queries & Masks & Avg. Segs. (A/V) \\
\midrule
Ref-YouTube-VOS & 3,978 & 5.45s  & 15,009 & 131k   & 1$^\dagger$ \\
MeViS           & 2,006 & 13.16s & 33,072 & 443k   & 1$^\dagger$ \\
Ref-DAVIS17     & 90    & 2.87s  & 1,544  & 13.5k  & 1$^\dagger$ \\
GroundMore      & 1.7k  & 9.61s  & 7.6k   & 249.5k & 1$^\dagger$ \\
\midrule
\bench          & 2,077 & \textbf{45.81s} & 16,250 & \textbf{1,569k} & \textbf{3.82 (2.18/1.64)} \\
\bottomrule
\end{tabular}}
\caption{
Statistics of representative language-guided video segmentation benchmarks.
A/V denotes active/vacant predicate-state segments.
$^\dagger$Existing datasets do not explicitly annotate active/vacant predicate states and are treated as single active segments for comparison.
}
\label{tab:stats}
\vspace{-14pt}
\end{table}

\begin{figure*}[t]
    \centering
    \includegraphics[width=1\linewidth]{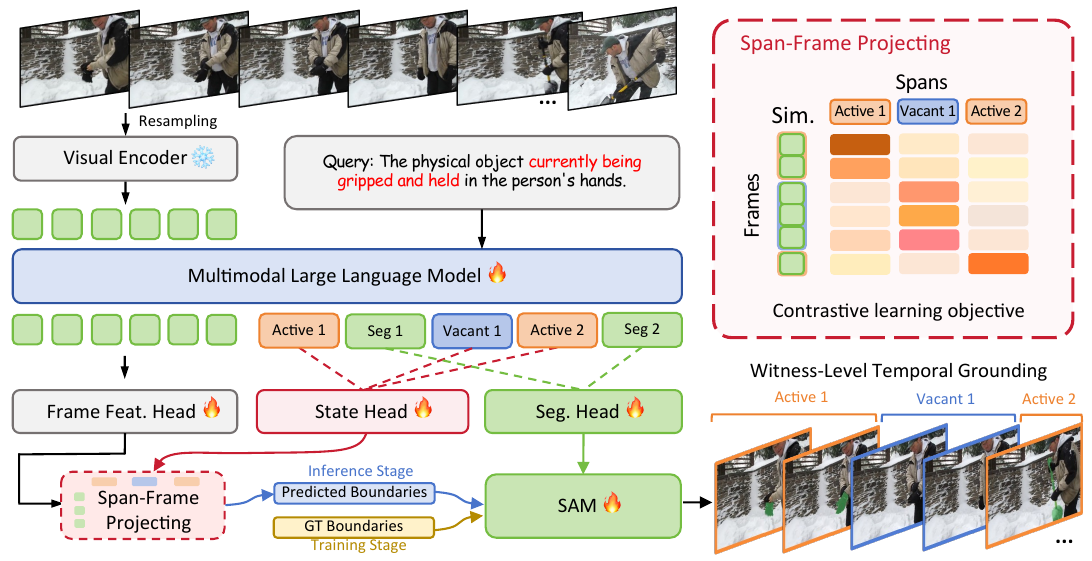}
    \caption{\textbf{Overview of the Predicate-State Grounding (PSG) Segmentation architecture.} PSG decouples semantic validity from mask generation via a strict \textit{State-Bind-Emit} pipeline. \textbf{(1) State:} A Multimodal Large Language Model (MLLM) infers the dynamic predicate states by generating discrete structural reasoning tokens (\texttt{[ACTIVE]}, \texttt{[VACANT]}, and \texttt{[SEG]}). \textbf{(2) Bind:} The Span-Frame Projecting module performs dense cross-modal alignment, binding these abstract state tokens to continuous video frames to establish frame-level temporal boundaries. \textbf{(3) Emit:} The \texttt{[SEG]} token acts as a semantic prompt for SAM to decode masks exclusively during active spans. Mask decoding is bypassed during predicted vacant intervals, substantially reducing invalid mask emission. }
    \label{fig:architecture}
\end{figure*}

\section{Predicate-State Video Reasoning Segmentation Framework}
\label{sec:task}

\subsection{Task Definition} 
\label{sec:task_definition}

Predicate-State Video Reasoning Segmentation (PS-VRS) takes as input a video clip $V \in \mathbb{R}^{T \times H \times W \times 3}$ and a natural language query $Q$, and outputs a sequence of temporal intervals with predicate states and pixel-level masks.
Unlike conventional Referring Video Object Segmentation (RVOS), which typically assumes an identity-consistent target track, PS-VRS treats the query as a time-dependent Boolean predicate $P_Q(o,t)\in\{0,1\}$ over objects and time.
The target is therefore the current visual witness satisfying the predicate, rather than a fixed object identity.
This requires the model to infer \textit{when} the predicate is active and \textit{what} should be segmented at each moment.

Formally, the output is defined as $\mathcal{O}=\{([t_s^i,t_e^i],s^i,\mathcal{M}^i)\}_{i=1}^{N}$, where $[t_s^i,t_e^i]$ denotes the temporal interval of the $i$-th semantic segment, and $s^i\in\{1,0\}$ denotes its predicate state.
Here, $s^i=1$ indicates an active state with a valid visual witness, while $s^i=0$ indicates a vacant state where no valid witness exists.
The spatial prediction $\mathcal{M}^i$ is conditioned on $s^i$:

{\small
\begin{equation}
    \mathcal{M}^i = 
    \begin{cases}
        \big\{ M_t \in \{0,1\}^{H \times W} \mid t_s^i \leq t \leq t_e^i \big\}, & s^i = 1, \\[1mm]
        \big\{ M_t = \emptyset \mid t_s^i \leq t \leq t_e^i \big\}, & s^i = 0.
    \end{cases}
\end{equation}
}

Therefore, active segments require pixel-level localization of the current visual witness, whereas vacant segments require the model to emit no mask.

\subsection{\bench Benchmark}

To support PS-VRS, we construct \bench, a benchmark for evaluating predicate-state reasoning, witness-level temporal grounding, and pixel-level segmentation in long-form videos.
We curate videos from diverse sources, including CoVLA~\citep{arai2025covlacomprehensivevisionlanguageactiondataset}, ActivityNet~\citep{caba2015activitynet}, and VDC~\citep{chai2025auroracapefficientperformantvideo}, whose multi-stage interactions and continuous scene evolution naturally induce predicates that may remain active, become vacant, or be re-instantiated by different visual witnesses.
Each query is annotated as predicate-state segments with active/vacant labels, target descriptions, reasoning annotations, and pixel-level masks for active intervals; dataset statistics are summarized in~\cref{tab:stats}. More detailed statistics on the dataset splits, source composition, and predicate-state distributions are provided in Appendix~\ref{app:data_annotation}.

\section{Method}

\subsection{Architecture}
\label{sec:method_architecture}

To address PS-VRS, we propose \method, a Predicate-State Grounding architecture built on an MLLM and SAM2.
As shown in~\cref{fig:architecture}, \method decomposes video reasoning segmentation into a \textbf{State--Bind--Emit} pipeline:
\begin{enumerate}[label=\arabic*), leftmargin=*, nosep]
    \item \textbf{State:} the MLLM predicts a predicate-state sequence with \texttt{[ACTIVE]} and \texttt{[VACANT]} tokens.
    \item \textbf{Bind:} the predicted state tokens are aligned to video frames to obtain temporal intervals for each predicate state.
    \item \textbf{Emit:} each active state is paired with a \texttt{[SEG]} token, whose hidden state is projected into a SAM2 prompt embedding for mask decoding; vacant states emit no mask.
\end{enumerate}
This decomposition separates semantic validity from mask generation, enabling different active segments to bind different visual witnesses while suppressing mask emission in vacant intervals.

\subsubsection{State Generation with Predicate-State Tokens}
\label{sec:state_generation}

Given a query $Q$ and sampled context frames $\{V_t\}$, the MLLM generates a discrete predicate-state sequence.
We use \texttt{[ACTIVE]} and \texttt{[VACANT]} to denote whether the query-defined predicate has a valid visual witness, and append a \texttt{[SEG]} token to each active state for mask emission.
For example, \texttt{[ACTIVE][SEG] [VACANT] [ACTIVE][SEG]} represents two active semantic segments separated by a vacant segment.
This sequence exposes predicate-state transitions before pixel-level mask decoding.

\subsubsection{Temporal Binding via State-Frame Alignment}
\label{sec:temporal_binding}

The generated predicate-state tokens specify the order of semantic segments, but they do not directly provide temporal boundaries.
To ground these discrete states in the video, we align state-token representations with frame-level visual features.

Specifically, let $\mathcal{H}=\{h_i\}_{i=1}^{N}\subset\mathbb{R}^{d_{\text{txt}}}$ denote the hidden states of the predicate-state tokens, and let $\mathcal{X}=\{x_t\}_{t=1}^{T}\subset\mathbb{R}^{d_{\text{vis}}}$ denote the visual features of sampled context frames.
We project them into a shared $D$-dimensional normalized space:
\begin{equation}
    v_i = \frac{\Phi_{\text{txt}}(h_i)}{\|\Phi_{\text{txt}}(h_i)\|_2}, 
    \quad 
    f_t = \frac{\Phi_{\text{vis}}(x_t)}{\|\Phi_{\text{vis}}(x_t)\|_2},
\end{equation}
where $\Phi_{\text{txt}}$ and $\Phi_{\text{vis}}$ are modality-specific MLPs.
We then compute a state-frame similarity matrix:
\begin{equation}
    S_{i,t} = \alpha \cdot \langle v_i, f_t \rangle,
\end{equation}
where $\alpha$ is a learnable logit scale.

During inference, we preserve the generated state order and locate each
boundary through adjacent-state competition.
We scan the sampled context frames temporally and transition from state
$i$ to state $i+1$ once the successor obtains higher similarity for
$w$ consecutive frames.
The resulting boundaries partition the video into ordered, contiguous,
and non-overlapping predicate-state intervals.

\subsubsection{State-Conditioned Mask Emission} 
For each grounded active segment ($s^i=1$), we extract the hidden state of its paired \texttt{[SEG]} token, $h_m^{\text{seg}}$, and project it into a visual prompt embedding.

Vacant segments ($s^i=0$) do not produce a \texttt{[SEG]} token and therefore do not invoke mask decoding.
Therefore, mask emission is conditioned on predicate state, allowing \method to generate masks only for active segments and reduce invalid predictions in vacant intervals.

\subsection{Joint Spatiotemporal Optimization}
\label{sec:optimization}

\nbf{Training data formulation.}
Since the MLLM processes videos as sampled context frames, we format each training example as a multi-image conversation.
Given uniformly sampled frames $\{V_1,\ldots,V_T\}$ and a query $Q$, the prompt is constructed as:

\begin{tcolorbox}[
    colback=gray!10,
    colframe=gray!40,
    boxrule=0.3pt,
    arc=2mm,
    left=2mm,
    right=2mm,
    top=1mm,
    bottom=1mm
]
\small
\textbf{USER:} 
\texttt{<image>$_1$ <image>$_2$ $\dots$ <image>$_T$ \{query\}} \\

\textbf{ASSISTANT:} 
\texttt{[ACTIVE][SEG][VACANT][ACTIVE][SEG]}
\end{tcolorbox}

The target token sequence is constructed from the ground-truth predicate-state segments:
\texttt{[ACTIVE]} and \texttt{[VACANT]} represent the ordered predicate states, while \texttt{[SEG]} is appended to each active state as the mask-emission token.
We optimize \method{} with a joint objective:
\begin{equation}
    \mathcal{L}
    =
    \lambda_{\mathrm{CE}}\mathcal{L}_{\mathrm{CE}}
    +
    \lambda_{\mathrm{temp}}\mathcal{L}_{\mathrm{temp}}
    +
    \lambda_{\mathrm{mask}}\mathcal{L}_{\mathrm{mask}}.
\end{equation}

\nbf{Language modeling loss.}
The language modeling loss supervises the autoregressive generation of the predicate-state sequence:
\begin{equation}
    \mathcal{L}_{\mathrm{CE}}
    =
    - \sum_{j=1}^{L}
    \log P(y_j \mid y_{<j}, V_{\mathrm{ctx}}, Q),
\end{equation}
where $y$ denotes the target token sequence consisting of \texttt{[ACTIVE]}, \texttt{[VACANT]}, and \texttt{[SEG]} tokens.
This loss trains the MLLM to represent predicate-state transitions in temporal order.

\nbf{State-frame alignment loss.}
To align predicate-state tokens with video frames, we supervise the state-frame similarity matrix $S$ using a binary alignment matrix $Y^{\mathrm{temp}}\in\{0,1\}^{N\times T}$.
Here, $Y^{\mathrm{temp}}_{i,t}=1$ if frame $t$ lies within the ground-truth temporal interval of semantic segment $i$, and $0$ otherwise.
We optimize the alignment with a BCE loss:
\begin{equation}
    \mathcal{L}_{\mathrm{temp}}
    =
    \frac{1}{NT}
    \sum_{i=1}^{N}
    \sum_{t=1}^{T}
    \mathcal{L}_{\mathrm{BCE}}
    \left(S_{i,t},Y^{\mathrm{temp}}_{i,t}\right).
\end{equation}
This objective provides temporal supervision for both active and vacant predicate states.

\nbf{Active-only mask supervision.}
Pixel-level supervision is applied only to active segments, where valid visual witnesses exist.
For each active segment $i$, we sample a denser set of supervision frames $\mathcal{F}^{\mathrm{act}}_i$ within its temporal interval and compute a BCE-Dice segmentation loss:
{\small
\begin{align}
    \mathcal{L}_{\mathrm{mask}}
    &=
    \frac{1}{\sum_{i:s^i=1} |\mathcal{F}^{\mathrm{act}}_i|}
    \sum_{\{i \mid s^i=1\}}
    \sum_{t\in\mathcal{F}^{\mathrm{act}}_i}
    \mathcal{L}_{\mathrm{seg}}^{(i,t)}, \\
    \mathcal{L}_{\mathrm{seg}}^{(i,t)}
    &=
    \mathcal{L}_{\mathrm{BCE}}
    \left(\hat{M}_t^{(i)},M_t^{(i)}\right)
    +
    \mathcal{L}_{\mathrm{Dice}}
    \left(\hat{M}_t^{(i)},M_t^{(i)}\right).
\end{align}
}
where $\hat{M}_t^{(i)}$ and $M_t^{(i)}$ denote the predicted and ground-truth masks.
By excluding vacant segments from mask supervision, the spatial objective is aligned with predicate-state validity.

\subsection{Inference}
\label{sec:inference}

During inference, \method samples context frames and uses the MLLM to generate an ordered predicate-state sequence.
We ground this sequence with localized adjacent-state competition, which assigns each context frame by comparing the similarity between its current state and the next state in the generated order.
A state transition is triggered only when the next state consistently obtains higher similarity within a sliding window of $w$ context frames, yielding stable and contiguous temporal intervals.
Unsampled dense frames inherit the assignment of their nearest context frame.

Mask decoding is invoked only for \texttt{[ACTIVE]} intervals.
For each active interval, the paired \texttt{[SEG]} token is projected into a SAM2 prompt embedding to decode masks within that interval, while \texttt{[VACANT]} intervals directly produce empty masks.
This segment-level inference confines mask generation to semantically valid intervals and reduces invalid emissions in vacant states.

\begin{table*}[!t]
\centering
\renewcommand{\arraystretch}{1.08}
\setlength{\tabcolsep}{5pt}

\resizebox{\textwidth}{!}{%
\begin{tabular}{l c ccc ccc ccc ccc}
\toprule

\multirow{3}{*}{\textbf{Method}}
& \multirow{3}{*}{\textbf{Params}}
& \multicolumn{9}{c}{\textbf{Spatial}}
& \multicolumn{3}{c}{\textbf{Temporal}} \\

\cmidrule(lr){3-11}
\cmidrule(lr){12-14}

&
& \multicolumn{3}{c}{$\mathcal{J}$}
& \multicolumn{3}{c}{$\mathcal{F}$}
& \multicolumn{3}{c}{$\mathcal{J}\&\mathcal{F}$}
& \multicolumn{3}{c}{$mIoU$} \\

\cmidrule(lr){3-5}
\cmidrule(lr){6-8}
\cmidrule(lr){9-11}
\cmidrule(lr){12-14}

&
& VT & TS & All
& VT & TS & All
& VT & TS & All
& VT & TS & All \\

\midrule

\rowcolor{gray!20}
\multicolumn{14}{c}{\textit{Predicate-State VRS Methods (Adapted from Large-scale MLLM)}} \\
\midrule

Gemini-3.1-Pro
& N/A
& 56.78 & 54.66 & 55.25
& 58.06 & 55.50 & 56.21
& 57.42 & 55.08 & 55.73
& 47.62 & 43.68 & 44.49 \\

Qwen3-VL-Max
& 235B
& 60.39 & 51.08 & 53.71
& 63.10 & 53.71 & 56.37
& 61.75 & 52.39 & 55.05
& 56.38 & 51.86 & 52.81 \\

Kimi-K2.5
& 1T
& 60.96 & 55.04 & 56.68
& 63.78 & 56.06 & 58.20
& 62.37 & 55.55 & 57.44
& 57.21 & 49.84 & 51.30 \\

Qwen3.5-Max
& N/A
& 63.84 & 54.15 & 56.87
& 68.49 & 56.97 & 60.20
& 66.17 & 55.56 & 58.54
& 62.56 & 57.04 & 58.21 \\

\midrule

\rowcolor{gray!20}
\multicolumn{14}{c}{\textit{Previous Video Reasoning Segmentation (VRS) Methods}} \\
\midrule

VISA
& 7B
& 30.91 & 20.19 & 23.15
& 34.14 & 23.70 & 26.59
& 32.53 & 21.94 & 24.87
& -- & -- & -- \\

VideoLISA
& 3.8B
& 34.04 & 25.40 & 27.78
& 36.65 & 28.02 & 30.41
& 35.34 & 26.71 & 29.09
& -- & -- & -- \\

VRS-HQ
& 13B
& 61.07 & 46.57 & 50.57
& 62.54 & 48.12 & 52.10
& 61.80 & 47.35 & 51.33
& -- & -- & -- \\

GLUS
& 7B
& 65.15 & 46.64 & 51.76
& 67.44 & 48.28 & 53.58
& 66.29 & 47.46 & 52.67
& -- & -- & -- \\

\midrule

\rowcolor{gray!20}
\multicolumn{14}{c}{\textit{Predicate-State VRS Methods (Base Model \& Ours)}} \\
\midrule

Qwen3.5-4B
& 4B
& 29.80 & 29.21 & 29.35
& 34.43 & 31.49 & 32.21
& 32.11 & 30.35 & 30.78
& 57.60 & 46.93 & 49.41 \\

\textbf{Ours}
& \textbf{4B}
& \textbf{65.58} & \textbf{59.61} & \textbf{61.26}
& \textbf{67.36} & \textbf{60.93} & \textbf{62.71}
& \textbf{66.47} & \textbf{60.27} & \textbf{61.99}
& \textbf{68.17} & \textbf{54.85} & \textbf{57.79} \\

\bottomrule
\end{tabular}%
}

\caption{\textbf{Quantitative comparison on \bench.}
The baseline methods are grouped into adapted large-scale MLLMs, previous VRS models, and our baseline comparison. Our proposed PSG establishes a new state-of-the-art for predicate-state VRS. It outperforms previous specialized methods by a large margin and demonstrates highly competitive spatial and temporal grounding against massive MLLMs using only 4B parameters. VT and TS denote the Visibility Transition and Target Switching subsets, respectively. "--" indicates the inability of previous VRS models to predict explicit temporal state boundaries, making temporal evaluation inapplicable. }
\label{tab:main_results}
\end{table*}

\section{Experiments}

\subsection{Experimental Settings}

\nbf{Implementation details.}
We build \method on Qwen3.5-4B~\citep{qwen3.5} with SAM2~\citep{ravi2024sam2segmentimages} as the mask decoder.
During training, we sample MLLM context frames uniformly at 1 fps, and sample denser mask-supervision frames at $2$ fps within active segments.
All experiments are conducted in bfloat16 precision with DeepSpeed ZeRO-2~\citep{DBLP:conf/sc/RajbhandariRRH20} on 4 NVIDIA H200 GPUs.
Additional optimization details are provided in~\cref{app:implementation_details}.

We train on a mixture of \bench, MeViS~\citep{mevis}, and Ref-YouTube-VOS~\citep{ref-youtube-vos}.
To make existing datasets compatible with PS-VRS, we convert their annotations into our predicate-state format: frames with valid masks are treated as active, while frames where the target disappears are marked as vacant.
Since these datasets are not designed for predicate-state reasoning, their converted samples mostly contain active segments, while \bench provides explicit active/vacant transitions and dynamic visual witnesses.

\nbf{Evaluation metrics.} 
We evaluate both spatial segmentation and temporal predicate-state grounding.
For spatial quality, we report the standard video object segmentation metrics \(\mathcal{J}\), \(\mathcal{F}\), and their mean \(\mathcal{J\&F}\).
For temporal grounding, we compute mIoU between predicted and ground-truth predicate-state intervals using greedy matching over continuous temporal IoU scores.

\nbf{Explicit state-segmented baselines.}
For generalist MLLMs that can reason over video but cannot directly output dense masks, we use a two-stage explicit baseline pipeline.
The model first outputs a structured temporal partition:
\begin{lstlisting}[basicstyle=\ttfamily\scriptsize, frame=single]
{
  "segments": [
    {
      "status": "active/vacant",
      "timerange": ["start", "end"],
      "bbox": [[x1, y1, x2, y2]] or null
    }
  ]
}
\end{lstlisting}
For active segments, the predicted boxes are used as prompts to initialize masks on key frames, followed by propagation within the predicted interval.
Vacant segments directly output empty masks.
The full prompt, parsing rules, and validity checks are provided in~\cref{app:explicit_baseline_prompt}.

\subsection{Main Results on \bench}
\label{sec:main_results}

We report the main results on \bench in~\cref{tab:main_results}.
We compare \method against large-scale MLLMs adapted with an explicit predicate-state pipeline, previous VRS models, and the Qwen3.5-4B base model.
Spatial quality is evaluated by \(\mathcal{J}\), \(\mathcal{F}\), and \(\mathcal{J\&F}\), while temporal predicate-state grounding is evaluated by mIoU.
Overall, \method achieves the best spatial performance and strong temporal grounding with a compact 4B backbone.

\nbf{Explicit predicate-state modeling enables temporal grounding.}
As shown in~\cref{tab:main_results}, previous VRS methods produce masks without explicit predicate-state intervals, so temporal mIoU is not directly applicable to them.
In contrast, \method explicitly predicts active/vacant temporal intervals and achieves $57.79$ overall mIoU, with $68.17$ mIoU on the VT subset.
Compared with the Qwen3.5-4B base model, \method improves overall mIoU from $49.41$ to $57.79$, showing that the proposed state-frame alignment is essential for grounding predicate states in time.

\nbf{Target-switching transitions are the main challenge.}
We split \bench into \textbf{Visibility Transition (VT)} and \textbf{Target Switching (TS)} subsets.
VT mainly involves changes in spatial presence, such as a target entering or leaving the camera view, while TS requires reasoning over query-conditioned state or identity changes.
TS accounts for $77.0\%$ of the transition cases and exposes a much larger gap between existing VRS methods and \method.
For example, GLUS~\citep{GLUS} drops from $66.29$ \(\mathcal{J\&F}\) on VT to $47.46$ on TS, whereas \method maintains $60.27$ \(\mathcal{J\&F}\) on TS, improving over GLUS by $12.81$ points.
This suggests that TS is not merely a visibility problem, but requires tracking whether the query-defined predicate is satisfied by a changing visual witness.

\nbf{State-conditioned binding improves spatial segmentation.}
Beyond prior VRS models, \method also outperforms large-scale two-stage MLLM baselines in spatial segmentation.
It achieves the best overall \(\mathcal{J\&F}\) score of $61.99$, surpassing Qwen3.5-Max ($58.54$), Kimi-K2.5 ($57.44$), and Qwen3-VL-Max ($55.05$), despite using a compact 4B backbone.
Large MLLMs remain competitive in temporal mIoU, but their decoupled text-to-box-to-mask pipeline can lose fine-grained visual information during mask generation.
By contrast, \method directly binds active predicate states to mask-emission tokens, improving spatial quality while retaining explicit temporal grounding.

\subsection{Generalization to MeViS}

We further evaluate \method on MeViS, where it achieves the best $\mathcal{J\&F}$ score of $57.1$, outperforming VideoSeg-R1 by $1.8$ points and GLUS-a by $5.8$ points, as shown in~\cref{tab:mevis_results}.
To assess cross-dataset generalization, we additionally train \method exclusively on \bench and directly evaluate it on MeViS without in-domain supervision.
Even under this setting, \method remains highly competitive, achieving $51.5$ $\mathcal{J\&F}$ and surpassing GLUS-a and VRS-HQ.
These results suggest that predicate-state modeling generalizes beyond \bench while preserving strong spatial segmentation performance.

\begin{table}[t]
\centering
\small
\resizebox{\linewidth}{!}{
\begin{tabular}{l c c c}
\toprule
\textbf{Method} & $\mathcal{J}$ & $\mathcal{F}$ & $\mathcal{J\& F}$ \\
\midrule
VideoGLaMM & 42.1 & 48.2 & 45.2 \\
VideoLISA & 41.3 & 47.6 & 44.4 \\
VISA & 41.8 & 47.1 & 44.5 \\
CoT-RVS & 42.7 & 49.1 & 45.9 \\
ViLLa & 46.6 & 52.3 & 49.4 \\
VRS-HQ & 48.0 & 53.7 & 50.9 \\
GLUS-s & 47.5 & 53.2 & 50.3 \\
GLUS-a & 48.5 & 54.2 & 51.3 \\
VideoSeg-R1& 52.7 & 57.8 & 55.3 \\
\midrule

\textbf{Ours}$^\dagger$ & 47.8 & 55.2 & 51.5 \\

\textbf{Ours} & \textbf{54.1} & \textbf{60.2} & \textbf{57.1} \\
\bottomrule
\end{tabular}}
\caption{
Quantitative comparison on \textbf{MeViS}.
$^\dagger$ denotes our model trained exclusively on PS-Bench and directly evaluated on MeViS without in-domain supervision.
With the complete training mixture, PSG achieves the best overall performance.
}
\label{tab:mevis_results}
\end{table}

\begin{table}[t]
\centering
\small
\renewcommand{\arraystretch}{1.1}
\setlength{\tabcolsep}{8pt}
\begin{tabular}{l ccc}

\toprule
\textbf{Method} & $\mathcal{J}$ & $\mathcal{F}$ & $\mathcal{J\& F}$ \\
\midrule
\textbf{with state}
& \textbf{61.26}
& \textbf{62.71}
& \textbf{61.99} \\
w/o vacant
& 55.08
& 56.87
& 55.97 \\
w/o state
& 30.99
& 32.60
& 31.79 \\
\bottomrule
\end{tabular}
\caption{
Ablation study on predicate-state modeling.
Removing explicit state modeling causes a large performance drop, while replacing \texttt{[VACANT]} with thresholding also degrades performance.
}
\label{tab:ablation_vacant}
\end{table}

\begin{figure*}[!t]
\centering
\includegraphics[width=1\linewidth]{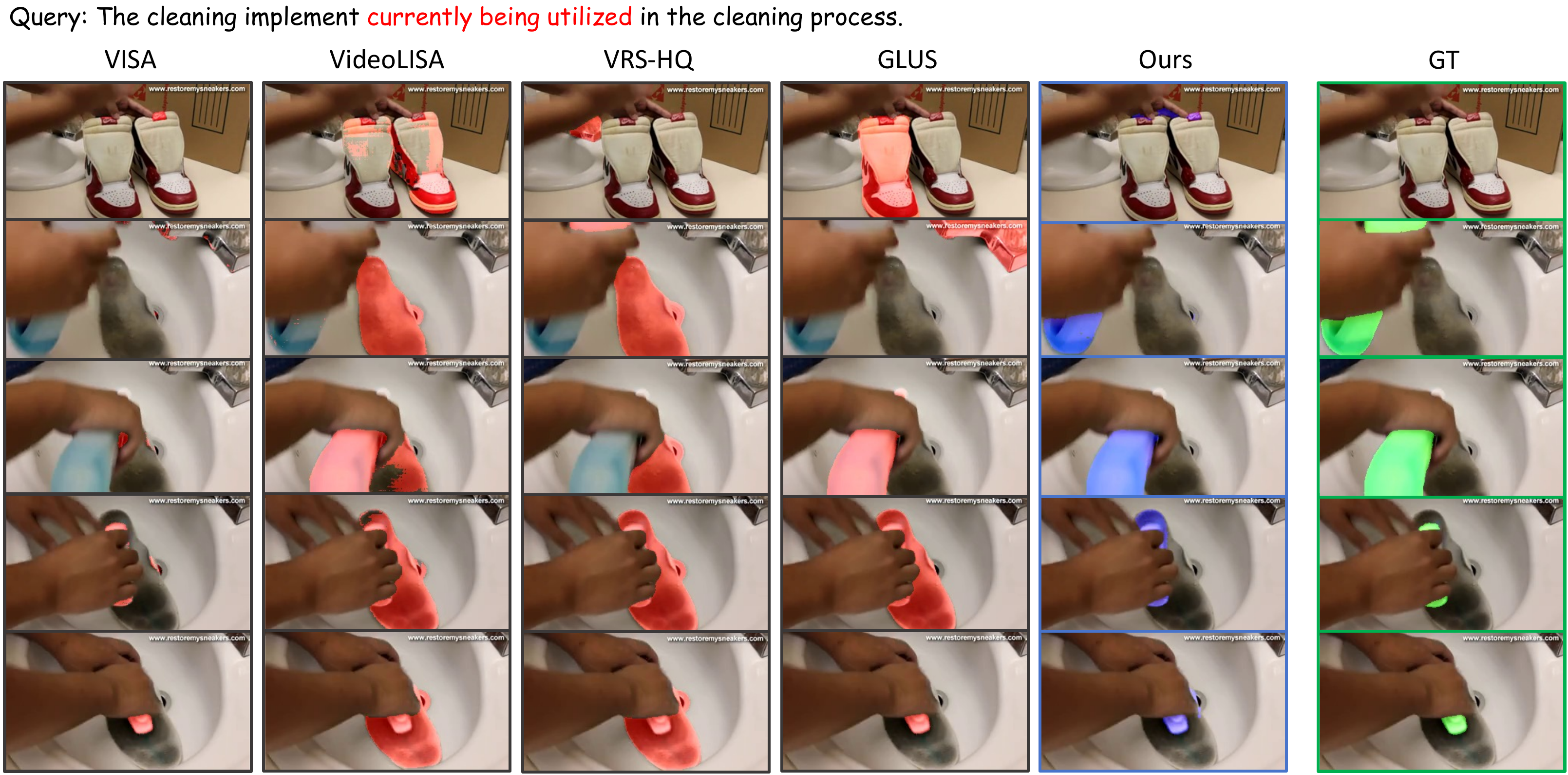}
\caption{
Qualitative comparison on dynamic predicate-state transitions.
For the query about the cleaning implement currently being used, the valid visual witness changes over time and is absent in the initial frames.
\method suppresses mask emission during the vacant state and rebinds the active state to the query-valid witness when the used implement changes.
In contrast, existing baselines often segment visually plausible but query-invalid objects, or continue emitting masks across invalid states.
}
\label{fig:visualization}
\end{figure*}

\begin{table}[t]
\centering
\small
\begin{tabular}{lcc}
\toprule
\textbf{Method}
& \textbf{False Active} $\downarrow$
& \textbf{False Vacant} $\downarrow$ \\
\midrule
VISA       & 49.23 & 1.13 \\
VideoLISA  & 53.67 & \textbf{0.12} \\
VRS-HQ     & 26.00 & 10.42 \\
GLUS       & 31.44 & 5.09 \\
\rowcolor{gray!20}
\textbf{Ours} & \textbf{19.70} & 12.09 \\
\bottomrule
\end{tabular}
\caption{
State errors on \bench.
False Active denotes the percentage of ground-truth vacant frames predicted as active, while False Vacant denotes the percentage of ground-truth active frames predicted as vacant.
}
\label{tab:state_error}
\end{table}

\subsection{Ablation Study}
\label{sec:ablation}

As shown in~\cref{tab:ablation_vacant}, we ablate the two key design choices in \method{}: explicit predicate-state modeling and explicit vacant-state modeling.

\nbf{Explicit predicate-state modeling is essential.}
Removing the state modeling branch causes a severe drop in $\mathcal{J\&F}$ from $61.99$ to $31.79$.
This variant directly decodes masks across frames without predicting active/vacant predicate states, indicating that spatial mask decoding alone cannot reliably determine when the query-defined predicate is valid.
Without an explicit temporal state variable, the model tends to treat every frame as a potential segmentation frame, conflating active targets with vacant or query-invalid intervals.

\nbf{Explicit \texttt{[VACANT]} modeling improves no-witness handling.}
Replacing \texttt{[VACANT]} with a fixed threshold reduces $\mathcal{J\&F}$ from $61.99$ to $55.97$.
Although thresholding can suppress some low-confidence predictions, it does not learn the semantic condition under which no valid witness exists.
This suggests that absence in PS-VRS is not merely a low-confidence case, but a query-conditioned semantic state that must be explicitly represented.
The learnable \texttt{[VACANT]} token better handles vacant or visible-but-invalid intervals, helping suppress invalid mask emission while preserving active-state segmentation quality.

\nbf{State error analysis.}
We further examine whether each frame is correctly identified as active or vacant. For methods without explicit state predictions, we treat a frame as
active when the model emits a non-empty mask and as vacant otherwise. As shown in~\cref{tab:state_error}, previous VRS methods tend to predict most frames as active, resulting in low False Vacant rates but frequent False Active errors.
In contrast, \method reduces the False Active rate to $19.70\%$, compared with $26.00\%$ for VRS-HQ and $31.44\%$ for GLUS.
These results demonstrate that explicit predicate-state modeling substantially reduces spurious predictions during vacant intervals, although balancing the two types of state errors remains an open challenge.

\subsection{Qualitative Results}
\label{sec:qualitative}

\cref{fig:visualization} shows a challenging example where the query-defined target is not a fixed object identity, but the cleaning implement currently being used.
At the beginning of the sequence, no valid visual witness exists, so the correct behavior is to emit no mask.
As the interaction evolves, the query-valid witness appears and later changes with the cleaning action.
\method follows this predicate-state transition by suppressing masks in the vacant state and decoding masks only for the active witness.
By contrast, prior VRS baselines often select visually salient but query-invalid objects, or keep emitting masks when the predicate state changes.
This example illustrates the benefit of explicit predicate-state grounding for both invalid-mask suppression and dynamic witness binding.

\section{Conclusion}

We introduced Predicate-State Video Reasoning Segmentation (PS-VRS), which formulates a language query as a predicate over objects and time rather than a fixed object identity.
To support this task, we constructed \bench with active/vacant predicate-state annotations, dynamic visual witnesses, temporal intervals, and pixel-level masks.
We further proposed \method, a Predicate-State Grounding method that separates state inference, witness binding, and state-conditioned mask emission through explicit \texttt{[ACTIVE]}, \texttt{[VACANT]}, and \texttt{[SEG]} tokens.
Experiments show that explicit predicate-state modeling improves full-video grounding, especially under query-conditioned transitions where existing methods often over-emit masks in invalid or vacant states.
We hope PS-VRS and \bench encourage future work on state-aware video reasoning segmentation beyond fixed-object grounding.

\section*{Limitations}

\method grounds predicate states from context frames sampled at 1 fps, with unsampled frames inheriting their nearest-frame assignments.
While interpretable, this design quantizes temporal boundaries and may miss rapid or ambiguous transitions.
Moreover, despite achieving $73.70\%$ greedy-match accuracy, its segment-count and exact-sequence accuracies are only $40.57\%$ and $30.50\%$, indicating that local matches do not guarantee recovery of the complete state structure.
Adaptive sampling, boundary refinement, and hierarchical state modeling may improve temporal precision and sequence consistency.

Our current implementation uses a compact 4B MLLM and SAM2 because long-video training and dense mask supervision are computationally demanding.
We leave scaling to larger reasoning backbones, longer video contexts, and stronger segmentation models for future work.

\section*{Acknowledgments}

This work was partially supported by the National Key Research and
Development Program of China under Grant No.~2024YFB2808903, the
Shenzhen Science and Technology Program under Grant
No.~ZDCY20250901103533010 and Tsinghua SIGS KA Cooperation Fund.
\bibliography{custom}

\clearpage
\appendix
\section*{Appendix}

\appendix

\section{Additional Implementation Details}
\label{app:implementation_details}

\nbf{Model architecture.}
We instantiate \method{} with Qwen3.5-4B~\citep{qwen3.5} as the multimodal large language model and SAM2~\citep{ravi2024sam2segmentimages} as the mask decoder.
The predicate-state grounding module contains modality-specific projection heads for predicate-state tokens and frame features, which are used to compute the state-frame similarity matrix.
The \texttt{[SEG]} hidden state is mapped to a SAM2 prompt embedding through an MLP prompt projector.

\nbf{Trainable parameters.}
We apply LoRA~\citep{lora} to the attention projections of the MLLM with rank $r=8$ and scaling factor $\alpha=16$.
The LoRA adapters, task-specific token embeddings, state-frame projection heads, prompt projection head, and SAM2 mask decoder are trainable.
The vision encoder is kept frozen to preserve its visual representation ability.

\nbf{Optimization.}
We use AdamW with $\beta_1=0.9$ and $\beta_2=0.95$.
The base learning rate is set to $3\times10^{-4}$ and follows a cosine decay schedule after a 10-step linear warmup.
We train the model for 30 epochs with gradient clipping set to 1.0.
Training is conducted in bfloat16 precision with DeepSpeed ZeRO-2~\citep{DBLP:conf/sc/RajbhandariRRH20} on 4 NVIDIA H200 GPUs.

\nbf{Frame sampling.}
For MLLM context input, videos are uniformly sampled at 1 fps and capped at a maximum of 120 context frames.
For mask supervision, we sample denser supervision frames at 2 fps within ground-truth active intervals and cap the total number of mask-supervision frames at 240.
At inference time, temporal grounding is performed on sampled context frames, and unsampled dense frames inherit the predicate-state assignment of their nearest sampled context frame.

\section{Training Data Conversion}
\label{app:data_conversion}

We train \method{} on a mixture of \bench, MeViS~\citep{mevis}, and Ref-YouTube-VOS~\citep{ref-youtube-vos}.
Since MeViS and Ref-YouTube-VOS are not originally annotated in the PS-VRS format, we convert their annotations into predicate-state sequences.

For each converted sample, frames with valid target masks are assigned to an active state and supervised with a \texttt{[ACTIVE][SEG]} token pair.
Frames in which the target is absent are assigned to a vacant state and supervised with a \texttt{[VACANT]} token.
Because these datasets are primarily designed for referring video segmentation, most converted examples contain a single active segment.
Nevertheless, the conversion allows them to provide useful spatial mask supervision while preserving vacant intervals when target disappearance is annotated.

\section{Explicit State-Segmented Baseline Pipeline}
\label{app:explicit_baseline_prompt}

Some generalist multimodal models can perform temporal reasoning but do not directly output dense masks.
To evaluate them under PS-VRS, we implement a decoupled two-stage explicit state-segmented pipeline.
The first stage asks the model to produce predicate-state intervals and representative spatial boxes.
The second stage uses the predicted boxes as prompts for mask generation and propagation.

\nbf{Prompt objective.}
The model is instructed to partition the entire video into continuous, non-overlapping temporal segments.
Each segment should be assigned either an \texttt{active} status, indicating that a valid visual witness satisfies the query-defined predicate, or a \texttt{vacant} status, indicating that no valid witness exists.
For each active segment, the model also predicts a representative timestamp and a bounding box for the visual witness.
Vacant segments contain no bounding box.

\nbf{Output schema.}
The model is required to output a JSON object in the following format:
\begin{lstlisting}[
    basicstyle=\ttfamily\tiny,
    frame=single,
    breaklines=true,
    breakatwhitespace=false,
    columns=fullflexible,
    xleftmargin=0.02\linewidth,
    xrightmargin=0.02\linewidth
]
{
  "segments": [
    {
      "clip_index": 0,
      "status": "active",
      "timerange": ["00:00.00", "00:05.00"],
      "target_description": "query-valid witness",
      "reasoning": "brief justification",
      "bbox": {
        "frame_timestamp": 2.50,
        "bbox_2d": [[x1, y1, x2, y2]]
      }
    },
    {
      "clip_index": 1,
      "status": "vacant",
      "timerange": ["00:05.00", "00:08.00"],
      "target_description": null,
      "reasoning": "no valid witness",
      "bbox": null
    }
  ]
}
\end{lstlisting}

\nbf{Validity checks.}
We parse the predicted JSON and sort segments by their starting times.
Invalid segments, malformed timestamps, or boxes outside the image boundary are discarded or clipped to valid ranges.
When the predicted segments do not fully cover the video, uncovered regions are assigned the nearest neighboring state to obtain a complete temporal partition.

\nbf{Mask generation and propagation.}
For each active segment, the predicted bounding box is used as a spatial prompt to initialize a mask on the representative key frame.
The initialized mask is then propagated within the predicted active temporal interval using the same video mask decoder.
For vacant segments, the output is directly set to an empty mask for all frames in the interval.
This pipeline enables generalist models to be evaluated under the PS-VRS protocol while keeping temporal reasoning and mask generation decoupled.

\begin{figure*}[t]
\centering
\begin{tcolorbox}[
    enhanced,
    colback=gray!8,
    colframe=gray!50,
    boxrule=0.5pt,
    arc=3pt,
    width=\textwidth,
    left=8pt,
    right=8pt,
    top=6pt,
    bottom=6pt,
    title=\textbf{System Prompt for Explicit Baseline Evaluation},
    coltitle=black,
    colbacktitle=gray!15,
    fonttitle=\bfseries\small,
    fontupper=\scriptsize,
    bottomtitle=3pt,
    toptitle=3pt
]

\textbf{\# Role.}
You are a Precision Video Reasoning Expert for query-conditioned temporal segmentation with possible multiple simultaneous targets.

\vspace{1mm}
\textbf{\# Task Objective.}
Given a video and a text query, segment the full video into sequential temporal clips and output strict JSON.
The query is fixed, but the physical targets satisfying it may change over time.
A new segment must begin whenever the target set satisfying the query changes.

\vspace{1mm}
\textbf{\# Core Segmentation Rules.}
\begin{enumerate}[label=\arabic*., leftmargin=*, itemsep=0pt, topsep=1pt, parsep=0pt]
    \item The output must cover the whole video from beginning to end.
    \item Segments must be sequential and non-overlapping.
    \item Keep one segment as long as the same target set continues to satisfy the query.
    \item Start a new segment whenever the target set changes, including target increase, target decrease, target replacement, vacant-to-active, or active-to-vacant transitions.
    \item Do not create unnecessary short segments; split only when the target set truly changes.
\end{enumerate}

\vspace{1mm}
\textbf{\# Status Rules.}
Each segment must have exactly one status:
\texttt{"active"} if one or more entities satisfy the query, and \texttt{"vacant"} otherwise.

\vspace{1mm}
\textbf{\# Vacant \& Active Segment Rules.}
If \texttt{status} is \texttt{"vacant"}, \texttt{bbox} must be \texttt{null}.
If \texttt{status} is \texttt{"active"}, all active targets must be localized in the same representative frame, and \texttt{bbox.bbox\_2d} must contain one box per active target.

\vspace{1mm}
\textbf{\# Multi-Target Bbox Rules.}
For one active target, \texttt{bbox.bbox\_2d} is a single box \texttt{[x1, y1, x2, y2]}.
For multiple active targets, it is a list of boxes from the same representative frame:
\texttt{[[x1, y1, x2, y2], [x1, y1, x2, y2], ...]}.

\vspace{1mm}
\textbf{\# Time \& Bounding Box Rules.}
\begin{enumerate}[label=\arabic*., leftmargin=*, itemsep=0pt, topsep=1pt, parsep=0pt]
    \item \texttt{timerange} must use \texttt{mm:ss.ss}, e.g., \texttt{["00:01.64", "00:06.32"]}.
    \item Segments must cover the whole timeline and connect cleanly without overlap.
    \item Bounding box coordinates must be relative coordinates in \texttt{[0, 1000]}.
    \item For active segments, choose one representative frame where all active targets are clearly visible.
    \item Do not output multiple \texttt{bbox} objects or duplicate \texttt{frame\_timestamp} fields for one segment.
\end{enumerate}

\vspace{1mm}
\textbf{\# Output Format.}
Output strict JSON only, without markdown or explanations:
{\ttfamily
\begin{quote}
\{
  "segments": [
    \{
      "clip\_index": 0,
      "status": "active",
      "timerange": ["mm:ss.ss", "mm:ss.ss"],
      "bbox": \{
        "frame\_timestamp": 0.0,
        "bbox\_2d": [x1, y1, x2, y2]
      \}
    \}
  ]
\}
\end{quote}
}
For vacant segments, use \texttt{"bbox": null}.
For multiple active targets, \texttt{bbox.bbox\_2d} must be a list of boxes.

\vspace{1mm}
\textbf{\# Final Checks.}
Before output, verify full video coverage, non-overlapping segments, valid \texttt{mm:ss.ss} timeranges, strict JSON format, and \texttt{bbox: null} for all vacant segments.

\end{tcolorbox}
\caption{Explicit baseline prompt.
System prompt used to adapt generalist MLLMs to predicate-state temporal segmentation with explicit active/vacant states, target-set transitions, and bounding-box localization.}
\label{fig:explicit_baseline_prompt}
\end{figure*}

\nbf{System prompt.}
The complete system prompt used for explicit temporal reasoning is shown in~\cref{fig:explicit_baseline_prompt}.

\section{Additional Evaluation Details}
\label{app:evaluation_details}

\nbf{Spatial metrics.}
We evaluate pixel-level segmentation using standard video object segmentation metrics.
Region similarity is measured by $\mathcal{J}$, contour accuracy by $\mathcal{F}$, and their mean by $\mathcal{J\&F}$.
For methods that do not output masks on vacant segments, empty masks are used for those intervals.

\nbf{Temporal grounding metric.}
Temporal predicate-state grounding is evaluated by matching predicted intervals to ground-truth predicate-state intervals.
We compute the temporal Intersection over Union (IoU) between predicted and ground-truth intervals and apply greedy matching within each video.
The reported mIoU is the average temporal IoU over matched intervals across active and vacant states.

\nbf{State error metrics.}
To diagnose predicate-state failures, we report two frame-level state error rates.
\emph{False Active} measures the percentage of ground-truth vacant frames predicted as active, reflecting invalid mask emission.
Conversely, \emph{False Vacant} measures the percentage of ground-truth active frames predicted as vacant, reflecting missed mask emission.
Together, they characterize the trade-off between over-prediction and over-suppression.

\nbf{Inference Efficiency.}
We compare the end-to-end inference efficiency of \method with representative VRS methods under the same hardware and implementation protocol. As shown in~\cref{tab:inference_efficiency}, \method is approximately $9\times$ faster than GLUS and $4\times$ faster than VideoLISA and VRS-HQ.
GLUS repeatedly performs multimodal reasoning over individual frames, while existing sequence-level methods still decode masks throughout the video.
In contrast, \method performs predicate-state reasoning over sparsely sampled context frames and invokes mask decoding only within predicted active intervals, avoiding redundant reasoning and segmentation during vacant periods.

\begin{table}[t]
\centering
\small
\begin{tabular}{lc}
\toprule
\textbf{Method}
& \textbf{Relative Runtime} $\downarrow$ \\
\midrule
VideoLISA & 4.0$\times$ \\
VRS-HQ    & 4.0$\times$ \\
GLUS      & 9.0$\times$ \\
\rowcolor{gray!20}
\textbf{Ours} & \textbf{1.0$\times$} \\
\bottomrule
\end{tabular}
\caption{
Normalized end-to-end inference time on the \bench test set, with the runtime of \method normalized to $1.0$.
All methods are evaluated under the same hardware and implementation protocol.
}
\label{tab:inference_efficiency}
\end{table}

\section{PS-Bench Construction and Statistics}
\label{app:data_annotation}
\nbf{Detailed Dataset Statistics.}
As summarized in~\cref{tab:dataset_statistics}, \bench follows a 7:2:1 train/validation/test split and contains videos from CoVLA~\citep{arai2025covlacomprehensivevisionlanguageactiondataset}, ActivityNet~\citep{caba2015activitynet}, and VDC~\citep{chai2025auroracapefficientperformantvideo}.
Active and vacant states account for $46.6\%$ and $53.4\%$ of the query-conditioned frame annotations, respectively, ensuring substantial coverage of no-witness periods.
Among the annotated transition cases, $22.96\%$ involve visibility transitions (VT), while $77.04\%$ involve target switching (TS), reflecting the benchmark's emphasis on dynamic visual witnesses.
\begin{table}[t]
\centering
\small
\begin{tabularx}{\linewidth}{@{}l >{\raggedright\arraybackslash}X@{}}
\toprule
\textbf{Category} & \textbf{Distribution} \\
\midrule
Data split
& Train / Val. / Test = 7:2:1 \\
Source videos
& CoVLA: 817; ActivityNet: 906; VDC: 354 \\
Predicate state
& Active: 46.6\%; Vacant: 53.4\% \\
Transition type
& VT: 22.96\%; TS: 77.04\% \\
\bottomrule
\end{tabularx}
\caption{
Detailed statistics of \bench.
The active/vacant distribution is computed over query-conditioned
frame annotations, while the VT/TS distribution is manually
computed over the test split.
}
\label{tab:dataset_statistics}
\end{table}

\nbf{Annotation Pipeline.} We construct \bench with a multi-stage annotation pipeline that combines multimodal model assistance, automatic mask propagation, and human verification.
SAM3 is used only for dataset annotation.

\begin{itemize}[leftmargin=*, itemsep=3pt, parsep=0pt]
    \item \textbf{Stage 1: Query and predicate-state generation.}
    We use Kimi-K2.5~\citep{kimiteam2026kimik25visualagentic} to generate reasoning-oriented queries with structured temporal annotations.
    Each query is decomposed into semantic segments with active/vacant states, temporal boundaries, target descriptions for active segments, and reasoning explanations.
    The generated queries are designed to cover predicate-state transitions, including target changes and vacant intervals.

    \item \textbf{Stage 2: Spatial grounding and human filtering.}
    For active segments, we use Qwen3-VL-32B-Instruct~\citep{bai2025qwen3vltechnicalreport} to localize representative visual witnesses with bounding boxes on selected key frames.
    Human annotators then refine temporal boundaries, check target consistency, and remove ambiguous or low-quality samples.

    \item \textbf{Stage 3: Dense mask propagation.}
    We use SAM3~\citep{carion2026sam3segmentconcepts} to propagate key-frame boxes into pixel-level masks within active temporal intervals.
    Vacant intervals are assigned empty masks, explicitly separating predicate-state reasoning from spatial mask generation.

    \item \textbf{Stage 4: Human verification.}
    We conduct an additional round of manual verification and mask refinement, with emphasis on the test set.
    This step improves spatial accuracy, temporal consistency, and semantic correctness across dynamic visual witnesses.
\end{itemize}

\nbf{Target-Switching Annotation.} A target-switching transition is annotated when the predicate remains active but its valid visual witness changes.
For example, given the query ``the nearest vehicle to the ego vehicle in the same lane,'' a change from vehicle A to vehicle B is represented as
$\mathrm{Active}(A)\rightarrow\mathrm{Active}(B)$,
without inserting an artificial vacant interval.

\nbf{Annotator qualification and briefing.}
The human annotation and verification process was conducted by a team of five qualified annotators, each holding at least a master's degree.
Annotators were instructed on the task definition, annotation protocol, quality requirements, and the intended research use of the data before participating in the annotation process.
The briefing covered active/vacant predicate-state labels, temporal boundary refinement, target consistency, and pixel-level mask verification.
Further personal or institutional details are omitted due to confidentiality agreements.

\begin{figure}[t]
    \centering
    \includegraphics[width=\columnwidth]
    {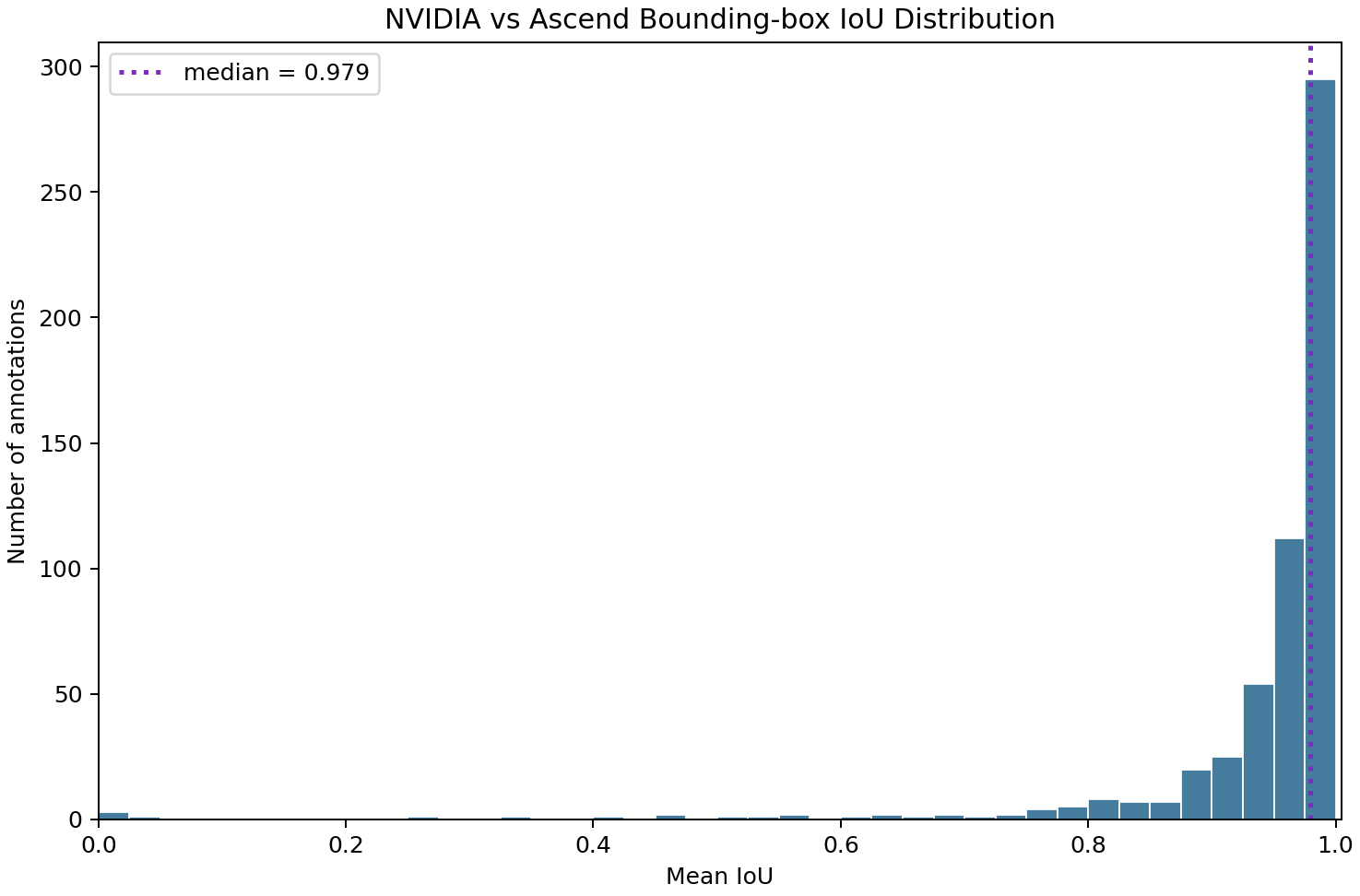}
    \caption{
    Distribution of pairwise bounding-box IoUs between the NVIDIA and
    Ascend deployments of Qwen3-VL-8B-Instruct.
    IoU is computed over the 559 annotations for which both deployments
    select the same candidate frame.
    The median IoU is $0.979$.
    }
    \label{fig:cross_platform_iou}
\end{figure}

\begin{figure}[t]
    \centering
    \includegraphics[
        width=\columnwidth,
        trim=28 12 28 12,
        clip
    ]{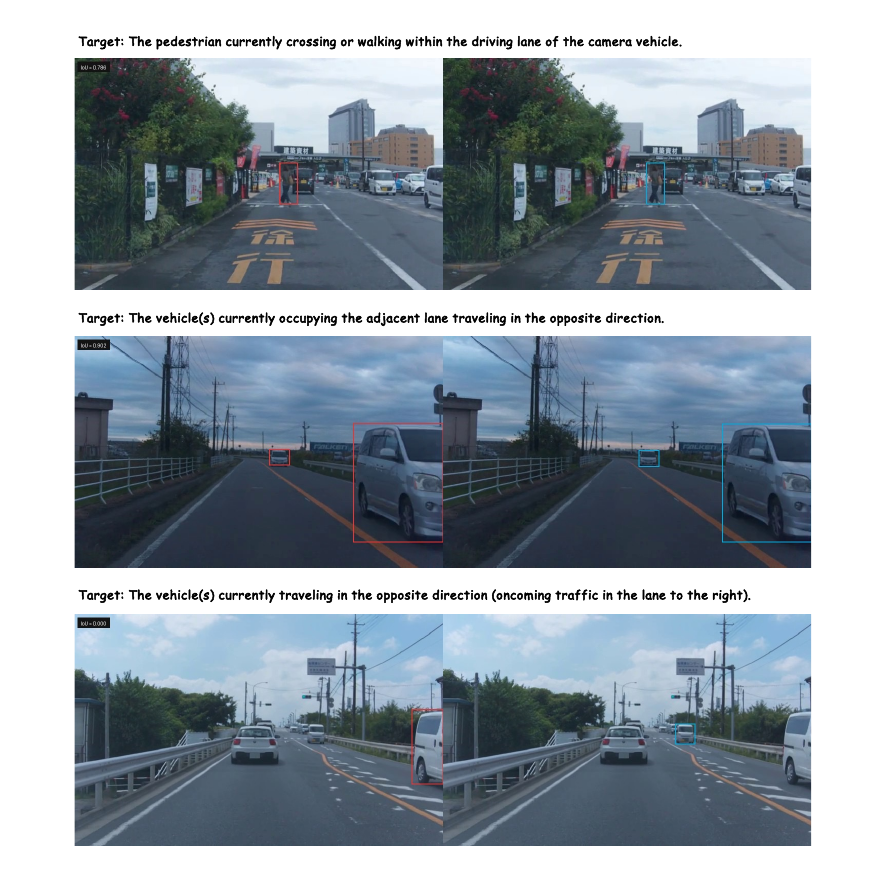}
    \caption{
    Qualitative cross-platform comparison of bounding-box annotations
generated by Qwen3-VL-8B-Instruct. Each row corresponds to the same
query and candidate key frame, with the NVIDIA prediction shown on the
left in red and the Ascend prediction shown on the right in cyan.
Pairwise IoU is reported in the upper-left corner. The examples include
both consistent and inconsistent predictions across the two deployment
stacks.
    }
    \label{fig:cross_platform_comparison}
\end{figure}

\nbf{Cross-platform consistency of spatial grounding.} We additionally conduct an auxiliary cross-platform study of the Stage~2
spatial-grounding procedure using Qwen3-VL-8B-Instruct. We deploy the
same model checkpoint on NVIDIA GPUs and Ascend NPUs and evaluate it
on 100 videos containing 942 segment annotations, including 641 active
segments. Both deployments use identical prompts, candidate key frames,
model weights, and greedy decoding with the temperature set to zero.
The valid-box generation rates on active segments are $88.61\%$ and
$88.77\%$ for NVIDIA and Ascend, respectively, while the bounding-box
presence agreement across all annotations reaches $99.68\%$. Among the
567 annotations for which both deployments produce a box, 559
($98.59\%$) select the same candidate frame. As shown in
Figure~\ref{fig:cross_platform_iou}, these same-frame predictions achieve
a mean IoU of 0.945 and a median IoU of 0.979. Moreover, $97.85\%$,
95.71\%, and 86.58\% of the paired predictions exceed IoU thresholds
of 0.5, 0.75, and 0.9, respectively, with 34.35\% producing exactly
identical boxes. Overall, the two deployment stacks exhibit highly
consistent spatial-grounding behavior, with no systematic degradation
observed on Ascend. As qualitatively illustrated in
Figure~\ref{fig:cross_platform_comparison}, the two deployment stacks
generally localize the same visual witnesses with closely aligned
bounding boxes, while occasional inconsistent predictions remain.

\section{Reproducibility and Release Plan}
\label{app:release_plan}

Upon publication, we will release the code, \bench annotations, benchmark splits, evaluation scripts, and baseline configurations to support reproducibility and future research on Predicate-State Video Reasoning Segmentation.
For videos derived from third-party sources, we will follow the corresponding dataset licenses and provide metadata or retrieval instructions when raw video redistribution is not permitted.
The code and evaluation toolkit will be released under the MIT License. As PSG can localize query-defined targets in videos, its deployment in privacy-sensitive applications should include appropriate privacy and data-use safeguards.

\end{document}